\documentclass[10pt,a4paper,twocolumn]{article}

\usepackage[left=2cm,right=2cm,top=2cm,bottom=2cm]{geometry}

\usepackage[utf8]{inputenc}
\usepackage[T1]{fontenc}
\usepackage[english]{babel}
\usepackage{newtxtext,newtxmath}
\usepackage{microtype}

\usepackage[hidelinks,implicit=false]{hyperref}
\usepackage{orcidlink}

\usepackage[citestyle=numeric-comp,bibstyle=numeric-comp,sorting=none,url=false,doi=false,useprefix=true,uniquename=false,uniquelist=false,giveninits=true,maxnames=2,natbib=true,backend=biber,hyperref=false]{biblatex}
\usepackage{csquotes}
\renewbibmacro{in:}{}

\DeclareFieldFormat[article,inbook,incollection,inproceedings,book,misc,unpublished]{title}{#1\isdot}
\DeclareNameAlias{default}{family-given}
\DeclareNameFormat{author}{%
    \ifnumequal{\value{uniquename}}{0}
    {\usebibmacro{name:family}
       {\namepartfamily}
       {\namepartgiven}
       {\namepartprefix}
       {\namepartsuffix}}
    {\usebibmacro{name:family-given}
       {\namepartfamily}
       {\namepartgiven}
       {\namepartprefix}
       {\namepartsuffix}}%
    \usebibmacro{name:andothers}}

\AtEveryBibitem{%
    \clearfield{volume}%
    \clearfield{number}%
    \clearfield{pages}%
}

\usepackage{makecell}
\usepackage{booktabs}
\usepackage{graphicx}
\usepackage{subcaption}
\usepackage[font=small,labelfont=bf,labelsep=period,singlelinecheck=false]{caption}

\usepackage{siunitx}
\DeclareSIUnit[quantity-product={}]{\percent}{\%}
\newcommand{\meansd}[3][2]{%
    $\num[round-mode=places,round-precision=#1]{#2} \pm %
    \num[round-mode=places,round-precision=#1]{#3}$%
}
\newcommand{\wilcoxon}[3]{%
    $W=\num[round-mode=places,round-precision=0,group-digits=none]{#1}$, %
    $p=\num[round-mode=places,round-precision=1,output-exponent-marker=e]{#2}$, %
    $\text{CLES}=\num[round-mode=places,round-precision=2]{#3}$%
}

\usepackage{datatool}
\DTLsettabseparator

\DTLloaddb{totals}{describe-totalcounts.tsv}
\DTLloaddb{toptags}{describe-toptags.tsv}
\DTLloaddb{topcategories}{describe-topcategories.tsv}
\DTLloaddb{demop}{describe-demographics_provided.tsv}
\DTLloaddb{clf}{validate-classifier_avg.tsv}
\DTLloaddb{liwcstats}{validate-liwc_stats.tsv}

\DTLassignfirstmatch{totals}{total}{posts}{\totalposts=count}
\DTLassignfirstmatch{totals}{total}{users}{\totalusers=count}

\DTLassignfirstmatch{demop}{demographic_variable}{gender}{\ngender=n_reported,\pctgender=pct_reported}
\DTLassignfirstmatch{demop}{demographic_variable}{age}{\nage=n_reported,\pctage=pct_reported}
\DTLassignfirstmatch{demop}{demographic_variable}{country}{\ncountry=n_reported,\pctcountry=pct_reported}
\DTLassignfirstmatch{demop}{demographic_variable}{gender+age+country}{\nall=n_reported,\pctall=pct_reported}

\DTLassignfirstmatch{clf}{scorer}{accuracy}{\clfaccm=cv_mean,\clfaccsd=cv_std}
\DTLassignfirstmatch{clf}{scorer}{f1}{\clffonem=cv_mean,\clffonesd=cv_std}

\DTLassignfirstmatch{liwcstats}{category}{insight}{\insightW=W_val,\insightP=p_val,\insightD=cohen-d,\insightCLES=CLES}
\DTLassignfirstmatch{liwcstats}{category}{agency}{\agencyW=W_val,\agencyP=p_val,\agencyD=cohen-d,\agencyCLES=CLES}

\title{\bf A large corpus of lucid and non-lucid dream reports}
\author{
    Remington Mallett~\orcidlink{0000-0001-6183-3098} \\ % chktex 8
    Department of Psychology \\
    Northwestern University \\
    \texttt{remington.mallett@u.northwestern.edu}
}
\date{}

\begin{document}
\begin{filecontents*}{describe-totalcounts.tsv}
total	count
posts	57778
users	5036
\end{filecontents*}

\begin{filecontents*}{describe-toptags.tsv}
tag	n_posts	n_users
lucid	1607	590
school	1550	429
flying	1211	389
family	1032	224
water	944	241
non-lucid	892	212
car	745	243
sex	613	185
fight	544	160
friends	537	208
house	536	202
false awakening	533	221
food	521	152
dream	502	268
music	466	168
driving	457	187
video game	453	88
work	452	111
dild	449	124
wild	413	177
death	405	226
mom	394	132
friend	378	117
dad	374	137
non - lucid	362	46
night	359	115
party	358	186
fighting	332	124
beach	331	163
dog	326	156
forest	323	136
ocean	319	139
home	312	104
bus	308	150
lucid dream	305	141
nightmare	302	204
game	301	107
city	292	122
cat	279	137
mall	279	110
zombies	277	142
dragon	274	91
fire	268	148
girl	265	140
hotel	263	137
bathroom	262	110
brother	262	104
mother	256	78
wife	253	27
train	250	128
weird	244	134
guns	237	120
college	235	82
sister	235	87
flight	233	42
snow	228	130
police	222	119
space	220	108
store	216	90
high school	214	76
animals	214	79
magic	213	119
pool	211	106
computer	206	93
travel	203	60
chase	198	88
dark	194	94
bed	193	85
restaurant	191	98
teacher	190	98
non lucid	189	89
totm	189	56
running	187	104
mirror	186	94
church	185	115
escape	185	79
telekinesis	185	65
vivid	184	126
money	183	103
gun	178	115
video games	177	76
war	175	109
river	173	83
deild	171	73
fragment	167	98
father	167	62
rain	165	106
demon	165	85
phone	164	96
love	164	107
cats	164	80
baby	163	75
fear	163	93
mild	163	39
hospital	158	92
boat	157	94
battle	157	64
minecraft	156	74
movie	155	96
swimming	154	97
harry potter	153	70
children	152	80
aliens	151	92
blood	151	90
lake	149	73
sleep paralysis	149	91
girlfriend	149	66
ghost	145	93
dogs	145	93
castle	145	95
building	145	52
bedroom	140	55
military	140	55
stairs	140	76
woman	140	39
monster	139	92
murder	139	87
reality check	138	69
elevator	136	79
zombie	136	91
darkness	135	70
scary	135	97
desert	134	84
celebrity	132	44
wbtb	132	70
class	131	72
girls	131	69
parents	131	64
television	130	79
fly	130	60
storm	126	72
drugs	126	78
obe	126	47
park	125	80
moon	121	77
mountain	119	73
vampires	119	40
kids	118	71
cave	117	78
competition	117	53
dreams	116	63
anime	114	41
pokemon	113	72
kiss	112	59
clothes	112	55
library	111	72
violence	110	47
toilet	110	50
naked	110	68
alcohol	109	61
white	108	36
lost	108	70
mansion	107	70
kitchen	107	60
fragments	106	69
memorable	106	61
apartment	106	60
ssild	105	23
airport	105	72
falling	105	63
time travel	105	65
island	105	67
portal	104	67
dream guide	104	70
alien	103	66
shopping	103	60
demons	103	43
vampire	102	63
truck	102	67
parkour	102	27
mountains	102	52
tree	101	65
ship	101	67
snake	100	66
sword	99	60
strange	99	70
red	99	52
monsters	98	48
cars	98	59
galantamine	98	11
fish	97	66
weapons	97	29
art	97	46
shooting	96	50
airplane	96	53
classroom	96	54
office	95	49
plane	94	63
flood	92	57
bridge	92	58
tornado	92	54
games	92	46
day	92	16
eating	91	44
dreamviews	91	33
university	91	41
bar	91	62
blue	91	48
wtf	91	26
time	90	44
book	90	50
transformation	90	36
classmates	90	27
bike	89	58
old friends	89	35
apocalypse	88	62
pain	88	60
singing	88	57
weed	87	51
climbing	87	37
sky	87	57
trees	87	46
healing	87	28
semi-lucid	87	68
stealing	87	38
shower	86	52
field	86	60
light	86	53
cops	86	46
dancing	85	50
dream journal	85	50
football	85	56
ice	84	55
sea	84	54
race	83	61
future	83	56
street	83	46
void	83	28
phasing	83	19
teleportation	82	48
star wars	82	48
anger	82	46
beer	82	33
old house	82	36
stars	80	49
dream fragment	80	43
room	80	48
bicycle	80	35
wedding	80	49
classmate	80	13
camera	79	43
window	79	48
guitar	79	54
evil	79	59
pizza	77	59
short	77	46
spider	76	59
books	76	36
door	76	49
knife	75	59
prison	75	55
clouds	74	50
vacation	74	51
road	74	44
robot	74	52
birds	74	51
motorcycle	74	52
wings	74	27
animal	73	28
baseball	73	33
danger	73	25
creepy	73	47
supernatural	73	26
basketball	72	41
floating	72	58
black	72	30
shop	71	43
helicopter	71	45
summoning	71	42
bomb	71	59
recall	71	27
shapeshifting	70	24
teleport	70	40
farm	69	32
concert	69	45
people	69	47
tower	69	50
cousin	69	45
child	68	40
garden	68	48
laptop	68	35
kissing	68	32
explosion	68	46
basement	68	48
ghosts	68	46
woods	68	52
man	68	41
bird	67	52
wind	67	39
town	66	43
japan	66	45
horse	66	49
jungle	65	52
cliff	64	42
walking	64	34
warehouse	63	43
boss	63	29
meditation	62	44
dance	62	50
crash	62	49
task of the month	62	30
theater	62	49
grandma	62	34
god	61	50
job	61	46
dream character	61	39
internet	60	41
scared	60	27
hair	60	43
old man	60	38
spaceship	60	42
wolf	60	41
bugs	60	37
green	60	32
maze	60	34
sun	60	43
boy	59	40
highway	59	48
temple	59	34
attack	59	35
christmas	58	47
underground	58	40
camp	58	44
snakes	58	42
cousins	58	21
shoes	58	44
funny	58	48
dreaming	58	36
boyfriend	58	37
japanese	58	26
gym	57	40
dying	57	31
lightning	57	43
band	57	41
lucid dreaming	57	39
fantasy	57	33
hands	56	32
doctor	56	36
combat	56	10
daughter	55	22
test	55	41
museum	55	29
candy	55	44
spiders	54	39
underwater	54	43
youtube	54	36
sleep	54	47
sad	54	39
coffee	54	34
killing	54	37
control	54	50
smoking	54	44
rescue	54	30
crime	54	34
boring	54	21
adventure	53	42
lion	53	35
marriage	53	29
army	53	43
holiday	53	16
grass	52	37
yellow	52	24
flowers	52	39
eyes	52	34
clarity	52	11
garage	52	32
crying	51	35
tunnel	51	40
action	51	26
dinosaur	50	35
batman	50	33
fun	50	40
robots	50	31
dorm	50	20
grocery store	50	31
hill	50	33
journal	49	35
blonde	49	13
fail	49	31
video	49	38
ruins	49	26
royalty	49	14
ufo	48	34
witch	48	36
earthquake	48	31
playground	48	39
hell	48	33
katana	48	11
ice cream	48	37
gold	48	29
birthday	48	42
factory	48	31
nonlucid	48	34
romance	48	32
awesome	48	28
supermarket	47	32
hiding	47	36
teeth	47	40
sand	47	39
subway	47	26
random	47	37
writing	47	28
parking lot	47	32
suicide	47	30
festival	47	34
religion	47	29
anxiety	47	22
possession	47	21
cake	47	39
grandparents	47	29
planet	47	28
cooking	46	35
doctor who	46	22
star trek	46	26
mission	46	21
inception	46	38
neighborhood	46	28
old friend	46	24
winter	45	34
dream control	45	35
dead	45	35
disney	45	21
accident	45	33
angel	45	29
doom	45	12
kid	44	24
glass	44	28
semi lucid	44	39
chocolate	44	39
wolves	44	31
crush	44	28
drunk	44	39
energy	44	30
cold	44	33
gods	44	15
bear	43	38
shark	43	32
van	43	34
giant	43	29
sunset	43	28
hallway	43	26
dragons	43	33
camping	43	33
village	43	30
teachers	43	26
trapped	42	34
theatre	42	20
mum	42	18
casino	42	29
late	42	28
cafeteria	42	31
roller coaster	42	31
epic	42	18
politics	42	26
long	42	34
son	42	19
cabin	41	30
dungeon	41	32
drowning	41	25
arcade	41	26
werewolf	41	29
story	41	18
china	41	29
cartoon	41	31
awareness	41	30
chased	40	24
drinking	40	21
club	40	32
mario	40	32
spirit	40	36
theft	40	26
relatives	40	14
clock	40	31
skyrim	40	24
rpg	40	22
grandmother	40	30
piano	40	30
3rd person	40	11
sunny	40	22
women	40	26
sexual	39	27
powers	39	24
cafe	39	29
ninja	39	29
backyard	39	28
haunted	39	27
creature	39	27
shadow	39	26
creatures	39	27
memory	39	27
sick	39	26
strangers	39	16
volcano	38	33
living room	38	21
fireworks	38	30
soccer	38	28
walmart	38	33
mirrors	38	12
table	38	25
end of the world	38	25
soldiers	38	25
survival	38	24
realistic	38	30
lava	37	28
witches	37	12
wine	37	24
horror	37	30
first lucid dream	37	37
paralysis	37	31
reading	37	22
confusion	37	31
painting	37	23
fence	37	28
taste	37	16
facebook	37	28
odd	37	22
wall	37	27
celebrities	37	19
king	36	31
fishing	36	23
run	36	25
assassin	36	22
shops	36	11
market	36	27
toty	36	16
marijuana	36	30
lights	36	23
movies	36	32
doors	36	25
haunted house	36	28
map	36	30
nap	36	28
medical	36	11
song	36	28
rape	36	32
grandma's house	36	10
treasure	35	27
old lady	35	31
coins	35	25
pirates	35	28
stadium	35	24
trains	35	26
highschool	35	18
little girl	35	19
first lucid	35	35
tv show	35	22
tiger	35	28
text	35	14
jumping	35	27
tattoo	35	24
car crash	35	29
notes	34	23
halloween	34	27
drawing	34	28
porn	34	23
trip	34	29
new york	34	24
swamp	34	28
lucidity	34	33
crowd	34	27
jesus	34	16
wood	34	25
kidnapping	34	27
amusement park	34	26
zelda	33	24
hometown	33	16
exam	33	21
hiking	33	16
horses	33	25
dinosaurs	33	28
tsunami	33	23
slide	33	27
training	33	21
aunt	33	26
portals	33	15
toys	33	23
vehicle	33	17
almost lucid	33	27
cop	33	28
past	33	25
contest	33	25
hug	33	23
puzzle	33	19
spy	33	23
tank	33	21
wild attempt	32	26
medieval	32	27
devil	32	21
caves	32	16
angry	32	27
carnival	32	27
dream recall	32	28
neighbor	32	23
swimming pool	32	25
sleeping	32	28
reflection	32	19
math	31	23
uncle	31	24
bank	31	26
earth	31	26
husband	31	14
futuristic	31	19
beautiful	30	25
destruction	30	21
jeep	30	14
students	30	21
homeless	30	24
childhood	30	21
meeting	30	21
experiment	30	24
spanish	30	23
attic	30	17
radio	30	28
bookstore	30	22
swords	30	21
spirits	30	15
cheating	30	26
power	30	23
melatonin	30	18
running away	30	22
being chased	30	13
thief	30	20
argument	30	27
pregnant	30	20
middle school	30	15
lab	30	14
sharks	29	24
killer	29	23
rocks	29	27
dream characters	29	23
couch	29	25
lord of the rings	29	18
sports	29	25
cookies	29	25
bees	29	24
gaming	29	12
show	29	22
rock	29	23
spinning	29	23
shared dream	29	21
ex girlfriend	29	10
pyramid	29	21
queen	29	26
dream sign	29	20
bully	29	22
deja vu	29	20
crazy	29	27
jump	29	24
nature	28	17
zombie apocalypse	28	18
gang	28	19
monkey	28	23
levitation	28	24
jail	28	25
gas station	28	25
puppy	28	22
wizard	28	20
exploration	28	13
roof	28	18
traveling	28	17
explosions	28	19
ride	28	18
stage	28	20
security	28	19
cage	28	21
smoke	28	24
hogwarts	28	20
houses	28	24
orange	28	22
halo	28	24
torture	28	25
martial arts	28	17
road trip	28	22
technology	27	18
robbery	27	24
chicken	27	22
weapon	27	17
waterfall	27	24
play	27	24
matrix	27	23
sadness	27	21
grandpa	27	25
poison	27	20
skyscraper	27	18
africa	27	22
purple	27	21
cards	27	15
cell phone	27	22
astral projection	27	21
vivid dream	27	26
panic	27	25
metal	27	20
abandoned	27	21
nudity	27	13
english	27	13
drive	27	18
elephant	27	23
mars	27	17
vibrations	27	16
kitten	27	18
machine	27	24
childhood home	27	10
asian	26	19
ladder	26	18
hero	26	23
kill	26	22
astral	26	12
lunch	26	19
film	26	20
gas	26	24
construction	26	18
babies	26	24
funeral	26	25
government	26	24
angels	26	20
california	26	17
colors	26	25
egypt	26	25
superpowers	26	19
goals	26	22
path	26	18
plants	26	21
train station	26	18
court	26	16
shotgun	26	17
group	26	20
world of warcraft	26	14
co-worker	26	12
naruto	25	21
numbers	25	21
aquarium	25	21
waves	25	21
guards	25	17
daytime	25	10
pond	25	20
golf	25	19
dinner	25	24
moving	25	23
pirate	25	20
happy	25	22
clothing	25	17
kittens	25	17
racing	25	23
buildings	25	20
flirting	25	18
theme park	25	20
telepathy	25	22
fog	25	22
journey	25	17
insect	25	18
shot	24	15
water park	24	21
breasts	24	10
glasses	24	22
annoying	24	21
stranger	24	20
undead	24	14
wow	24	14
videogames	24	11
invasion	24	11
investigation	24	15
morning	24	16
outer space	24	14
language	24	12
dream within a dream	24	23
lego	24	21
autumn	24	10
underwear	24	18
psychic	24	15
jogging	24	16
weather	24	18
confused	24	22
grey	24	15
princess	24	20
closet	23	17
heaven	23	18
ball	23	20
superman	23	22
alarm	23	21
fall	23	18
avatar	23	22
keys	23	22
boats	23	21
mud	23	19
shared	23	15
high	23	15
couple	23	19
eggs	23	20
gravity	23	23
my house	23	17
emotional	23	21
grandfather	23	18
nude	23	16
electricity	23	17
glitch	23	10
barn	23	21
bullies	23	11
twins	23	18
skeleton	23	19
arena	23	19
lady	23	16
mexico	23	16
shopping mall	23	13
windows	23	19
fps	22	10
emotions	22	18
thugs	22	14
worms	22	22
cleaning	22	19
paper	22	17
surgery	22	21
talking	22	19
search	22	17
school bus	22	17
auditorium	22	17
turtle	22	19
videogame	22	15
trailer	22	17
mask	22	16
cigarettes	22	16
pets	22	19
vision	22	22
movie theater	22	21
gore	22	20
traffic	22	19
embarrassment	22	13
hole	22	20
team	22	13
mafia	22	15
planets	22	17
teleporting	22	20
ipod	22	18
boxing	22	12
victorian	22	12
ants	22	17
fingers	22	16
fair	22	14
sisters	21	15
field trip	21	15
mcdonalds	21	18
detective	21	18
hunting	21	19
soldier	21	16
pee	21	17
doll	21	16
india	21	15
elementary school	21	18
london	21	15
simpsons	21	15
mystery	21	20
balcony	21	19
box	21	21
fox	21	15
wrestling	21	13
obama	21	18
america	21	11
blond	21	11
cliffs	21	14
conversation	21	16
memories	21	15
rabbit	21	20
tea	21	17
cool	21	16
amazing	21	19
science	21	18
dress	21	18
photos	21	11
tent	21	14
bowling	21	13
depression	21	14
virus	21	16
resort	21	16
brain	21	19
palace	21	14
super powers	21	14
italy	21	17
french	21	20
cult	21	17
rocket	21	12
dimension	21	19
bunker	21	12
pictures	21	18
emotion	21	13
space ship	20	16
graveyard	20	15
chasing	20	15
manga	20	14
skype	20	13
balloons	20	15
skiing	20	20
motorbike	20	14
zoo	20	18
skating	20	15
surreal	20	19
rollercoaster	20	15
dream signs	20	12
call of duty	20	15
biology	20	10
laser	20	19
hand	20	17
lol	20	14
spider-man	20	11
summon	20	14
date	20	16
computers	20	16
agent	20	13
my little pony	20	13
restroom	20	11
exploring	20	11
key	20	17
voice	20	20
priest	20	14
monk	20	14
lucid dreams	20	16
insomnia	20	11
kidnap	20	11
reality checks	20	17
ring	20	17
stalker	20	12
gross	20	18
armor	20	13
frog	20	15
teaching	20	14
awkward	20	18
stabilization	20	16
werewolves	20	13
satan	20	15
bread	19	19
searching	19	17
sound	19	15
staircase	19	11
virtual reality	19	19
stress	19	17
dream fragments	19	12
nuke	19	13
warrior	19	15
wand	19	16
superhero	19	18
subconscious	19	17
tanks	19	10
president	19	18
nightmares	19	19
axe	19	14
bug	19	14
meteor	19	14
dock	19	18
confusing	19	18
breakfast	19	19
ceiling	19	12
guide	19	17
disease	19	16
illness	19	12
betrayal	19	17
statue	19	15
sci-fi	19	16
mantra	19	12
sexy	19	11
chinese	19	14
pills	19	15
freeway	19	13
non-lucid dream	19	16
station	19	12
nurse	19	15
rainbow	19	15
old woman	19	11
knives	19	16
jealousy	19	14
health	19	14
execution	19	17
dentist	19	19
rats	19	15
guilt	19	17
hovering	19	13
olympics	19	14
germany	19	15
hallucination	18	14
bath	18	16
cruise ship	18	14
insects	18	14
iphone	18	15
plane crash	18	17
screaming	18	12
paris	18	14
injury	18	16
homestuck	18	11
hot	18	16
false	18	15
parking	18	14
burning	18	17
body	18	14
florida	18	14
bible	18	13
blurry	18	16
fortress	18	12
pregnancy	18	16
german	18	11
neighbors	18	14
invisible	18	15
crystals	18	14
butterfly	18	16
escalator	18	13
link	18	10
downtown	18	13
ex-girlfriend	18	14
false awakenings	18	18
car accident	18	15
hostage	18	16
pub	18	15
peeing	18	12
pet	18	15
success	18	12
apple	18	16
security guard	18	10
cancer	18	16
bikes	18	11
elements	18	13
pistol	18	11
knight	17	14
jewelry	17	14
trampoline	17	16
skeletons	17	13
mine	17	16
professor	17	13
toy	17	12
skateboard	17	15
post-apocalyptic	17	12
heart	17	17
recurring	17	16
bathtub	17	13
dirt	17	13
comics	17	11
canada	17	12
las vegas	17	12
milk	17	17
penis	17	11
hat	17	15
drink	17	17
airplanes	17	13
elevators	17	12
convention	17	12
walk	17	14
colosseum	17	11
interview	17	16
picture	17	15
car chase	17	16
revenge	17	14
scientist	17	15
world	17	17
gay	17	16
multiple dreams	17	17
lions	17	14
parade	17	16
quest	17	15
color	17	16
voices	17	13
egg	17	17
summer	17	14
owl	17	16
eye	17	15
bite	17	12
medicine	17	10
life	17	14
laundry	17	15
hide	17	15
soda	17	15
cigarette	17	16
cinema	17	11
shirt	17	16
nighttime	17	16
nazi	17	13
change	17	15
texting	17	17
female	17	11
sewer	17	16
fild	17	13
poker	16	14
hawaii	16	16
kidnapped	16	13
surfing	16	14
erotic	16	12
goddess	16	11
sandwich	16	15
shadows	16	14
cheese	16	16
sniper	16	15
superheroes	16	12
chicago	16	16
drums	16	14
swim	16	14
project	16	13
statues	16	10
mob	16	15
sonic	16	11
birth	16	14
acid	16	13
stealth	16	13
chair	16	12
intruder	16	11
frustration	16	14
rich	16	14
waiting	16	14
lizard	16	14
tribe	16	12
tornadoes	16	12
mcdonald's	16	12
speech	16	14
mermaid	16	16
hypnagogic hallucinations	16	11
tired	16	13
poop	16	14
prisoner	16	14
super strength	16	10
grenade	16	15
stuck	16	12
fallout	16	12
intense	15	15
dcs	15	12
pink	15	12
rifle	15	11
haircut	15	14
acting	15	14
dolphin	15	13
russia	15	11
sushi	15	13
thunderstorm	15	10
long dream	15	13
criminal	15	13
dark room	15	11
fbi	15	10
choir	15	12
pony	15	11
valley	15	13
trouble	15	14
alligator	15	12
futurama	15	11
department store	15	14
floor	15	13
crow	15	11
punch	15	15
vitamin b6	15	11
mind control	15	13
chat	15	10
tunnels	15	12
men	15	11
news	15	14
rave	15	11
sunlight	15	13
crystal	15	13
trash	15	13
trucks	15	12
star	15	11
newspaper	15	10
ninjas	15	15
goal	15	13
failure	15	11
watch	15	14
hunger games	15	13
forum	15	11
spear	15	11
grandparents' house	15	10
boxes	15	12
conspiracy	15	15
paint	15	13
bus stop	15	12
interesting	15	10
phone call	15	13
out of body experience	15	13
motel	15	13
stabilize	15	14
carpet	15	11
hallucinations	15	10
beard	14	13
fruit	14	13
skateboarding	14	12
cookie	14	13
lecture	14	14
walls	14	11
attacked	14	14
planes	14	14
calm	14	11
smell	14	13
graffiti	14	11
western	14	13
pants	14	13
towel	14	13
pillow	14	13
hockey	14	12
wallet	14	13
leaves	14	13
outdoors	14	12
photograph	14	10
awakening	14	12
real	14	12
parachute	14	12
face	14	13
galaxy	14	14
mushrooms	14	13
balloon	14	12
missile	14	10
homework	14	13
peace	14	13
gta	14	12
sickness	14	10
bee	14	12
bat	14	12
scream	14	14
stabbing	14	12
air	14	12
cavern	14	11
history	14	13
blind	14	11
gift	14	11
silver	14	10
flying car	14	10
business	14	10
fountain	14	13
mouse	14	14
chemistry	14	12
radiation	14	10
cloud	14	12
guy	14	12
maths	14	10
doctors	14	12
prince	14	13
coffin	14	12
happiness	14	12
costume	14	12
birthday party	14	14
gate	14	12
laboratory	14	13
disaster	14	11
short dream	14	13
heist	14	10
secret	14	11
cloudy	14	10
bears	14	12
wierd	14	13
resident evil	14	10
europe	14	13
final fantasy	13	11
snowboarding	13	12
poetry	13	10
crocodile	13	11
choking	13	12
photo	13	13
police officer	13	13
breathing	13	12
suit	13	12
goblin	13	12
morphing	13	10
advice	13	13
platform	13	13
symbol	13	12
samurai	13	13
shared dreaming	13	11
physics	13	12
wilderness	13	10
message	13	12
whale	13	13
refrigerator	13	11
best friend	13	12
rope	13	11
junkyard	13	10
rage	13	12
laughing	13	13
france	13	11
reality	13	12
costumes	13	11
bombs	13	11
skull	13	11
black cat	13	10
yoga	13	13
ipad	13	12
cathedral	13	10
head	13	12
deer	13	10
dating	13	12
broken	13	13
ex-boyfriend	13	10
indian	13	12
strange dream	13	11
telekenesis	13	12
boys	13	12
family guy	13	11
pit	13	12
deild attempt	13	11
astronaut	13	12
cruise	13	12
scorpion	13	13
raining	13	11
mice	13	11
terrorist	13	11
bright	13	11
non- lucid	13	10
terrorists	13	11
poor	13	11
nervous	13	11
rooftop	13	11
terminator	12	11
condom	12	10
turtles	12	11
slow motion	12	12
octopus	12	10
chickens	12	10
socks	12	10
challenge	12	12
labyrinth	12	10
waterbending	12	10
wal-mart	12	10
microwave	12	10
chess	12	11
furniture	12	11
lottery	12	10
gifts	12	10
alex	12	11
clown	12	11
commercial	12	12
fairy	12	11
pier	12	11
shaman	12	11
presence	12	10
corpse	12	11
shooter	12	10
blue eyes	12	10
awake	12	12
tickets	12	12
flower	12	11
black hole	12	12
bench	12	10
fridge	12	10
chainsaw	12	10
first ld	12	12
reality check fail	12	12
candle	12	10
usa	12	11
tour	12	12
jacket	12	11
attempt	12	11
xbox	12	12
whales	12	11
monkeys	12	11
alligators	12	11
peanut butter	12	11
circus	12	11
heat	12	10
tools	12	12
backpack	12	12
cyborg	12	11
pig	12	11
spell	12	11
dresses	12	11
wormhole	12	12
super speed	12	11
locker	11	10
ritual	11	10
mexican	11	10
gorilla	11	10
banana	11	10
tomb	11	10
dimensions	11	10
deild chain	11	10
wheelchair	11	11
lamp	11	10
irc	11	10
hallways	11	11
goat	11	11
deck	11	10
luggage	11	10
enemy	11	10
graduation	11	11
wasteland	11	11
arrested	11	10
bullying	11	10
mess	11	10
architecture	11	10
eclipse	11	11
performance	11	10
boots	11	10
gangs	11	11
giants	11	11
research	11	10
ticket	11	11
hotel room	11	11
vomit	11	11
kung fu	11	10
first time	11	10
apocalyptic	11	11
fish tank	11	10
locker room	11	10
stone	11	11
student	11	10
south park	11	10
rice	11	10
introduction	11	11
packing	11	11
new zealand	11	10
sinking	11	11
line	11	11
sweets	11	11
note	11	10
third person	11	10
pigs	11	11
diving	11	10
oil	10	10
bending	10	10
sheep	10	10
tropical	10	10
trap	10	10
annoyed	10	10
sniper rifle	10	10
raccoon	10	10
light switch	10	10
secret agent	10	10
sign	10	10
diner	10	10
prophecy	10	10
card	10	10
hills	10	10
bizarre	10	10
levitate	10	10
waking	10	10
harbor	10	10
james bond	10	10
umbrella	10	10
meat	10	10
excited	10	10
spiritual	10	10
bunny	10	10
gliding	10	10
arrows	10	10
villain	10	10
twitter	10	10
comic	10	10
lesson	10	10
tentacles	10	10
speed	10	10
levitating	10	10
psychedelic	10	10
arrow	10	10
tavern	10	10
obstacle course	10	10
\end{filecontents*}

\begin{filecontents*}{describe-topcategories.tsv}
category	n_posts	n_users
non-lucid	31941	2967
lucid	14342	2431
uncategorized	13028	2439
dream fragment	9448	1690
memorable	8325	2068
side notes	3053	596
false awakening	2930	989
nightmare	2483	1086
task of the month	935	296
task of the year	270	109
\end{filecontents*}

\begin{filecontents*}{describe-demographics_provided.tsv}
demographic_variable	n_reported	pct_reported
gender	3169	63
age	577	11
country	3509	70
gender+age	550	11
gender+country	2520	50
age+country	459	9
gender+age+country	447	9
\end{filecontents*}

\begin{filecontents*}{validate-classifier_avg.tsv}
scorer	cv_mean	cv_std
accuracy	0.76974	0.02050
f1	0.76522	0.02326
precision	0.78021	0.01912
recall	0.75107	0.03093
roc_auc	0.76973	0.02053
\end{filecontents*}

\begin{filecontents*}{validate-liwc_stats.tsv}
category	W_val	p_val	RBC	CLES	cohen-d	cohen-d_lo	cohen-d_hi	n
insight	212928.00000	1.62781e-33	0.40307	0.62192	0.40167	0.32831	0.46922	1195
agency	166811.00000	2.27027e-57	0.53314	0.67572	0.55945	0.46699	0.64222	1195
\end{filecontents*}

\maketitle

\section*{Abstract}
All varieties of dreaming remain a mystery. Lucid dreams in particular, or those characterized by awareness of the dream, are notoriously difficult to study. Their scarce prevalence and resistance to deliberate induction make it difficult to obtain a sizeable corpus of lucid dream reports. The consequent lack of clarity around lucid dream phenomenology has left the many purported applications of lucidity under-realized. Here, a large corpus of 55k dream reports from 5k contributors is curated, described, and validated for future research. Ten years of publicly available dream reports were scraped from an online forum where users share anonymous dream journals. Importantly, users optionally categorize their dream as lucid, non-lucid, or a nightmare, offering a user-provided labeling system that includes 10k lucid and 25k non-lucid, and 2k nightmare labels. After characterizing the corpus with descriptive statistics and visualizations, construct validation shows that language patterns in lucid-labeled reports are consistent with known characteristics of lucid dreams. While the entire corpus has broad value for dream science, the labeled subset is particularly powerful for new discoveries in lucid dream studies.

\bigskip
\noindent {\bf Keywords:} \emph{sleep}, \emph{dreaming}, \emph{natural language processing}

\section{Introduction}\label{sec:introduction}

Lucid dreams are those that include awareness of the dream while dreaming~\cite{baird2019}. They are often associated with control over dream content~\cite{lemyre2020}, which is commonly used to experience fun and impossible behaviors while sleeping~\cite{stumbrys2014}. Despite early promise of lucid dreaming applications in therapy~\cite{demacedo2019} and research~\cite{konkoly2021}, the variety and detail of what it is \emph{like} to be lucid while dreaming is still unclear~\cite{mallett2021}. This conflict, of application endorsement on the one hand and limited scientific understanding on the other, has raised concerns about whether some lucid dreaming practices are being implemented too soon~\cite{vallat2019,soffer2020}. Are we ``leaping before we look'' with respect to lucid dreaming applications?

\par
Without sufficient data on lucid dreams, it is difficult to ``look'' at all. Quantitative studies of lucid dreaming are often hindered by low sample sizes. Lucid dream occurrence is relatively rare in nature~\cite{saunders2016} and they are difficult to induce reliably~\cite{stumbrys2012}, making it difficult to accumulate enough data for strong conclusions. To overcome this limitation, the current project takes advantage of a rising trend in computational social science, the use of digital trace data to accumulate large and naturalistic datasets~\cite{lazer2021}.

\par
The ubiquity of digital media in daily life has provided a major data source for psychologists~\cite{lazer2021,shah2015,kern2016,gosling2015,rafaeli2019,heng2018}. The ability to view online human behavior en masse has been hailed as social science's first true ``telescope''~\cite{jungherr2017}. Datasets curated from popular social media sites (e.g., Twitter, Reddit) are not only large, but also free from experimenter intervention or influence, a major methodological concern in psychological research that dream studies are not immune to~\cite{picard2021, schredl2002}. Thus, dream science could benefit as well from the use of digital trace data~\cite{bulkeley2017}, and indeed has in recent years~\cite{sanz2018, niederhoffer2017, elce2021}. While dream reports are harder to extract from mainstream social media sites, it is common practice for researchers to scrape specialized data from targeted forums~\cite{speckmann2021}. Here, this approach was taken to build a large corpus of (lucid) dreams from DreamViews\footnote{https://www.dreamviews.com}, an online lucid dreaming journal active since 2010.

\par
The current paper presents the curation, characterization, and validation of the DreamViews corpus. First, the process of data collection, cleaning, and label extraction (lucid, non-lucid, and nightmare) is described. Second, the general size and details of the final corpus are presented. Last, natural language processing tools~\cite{hirschberg2015,elce2021} are used to show that the lucid, non-lucid, and nightmare labels provided by users have construct validity. That is, the content of lucid dreams and nightmares in the corpus show expected properties based on prior literature~\cite{voss2013,paquet2020} and justify the use of these labels in future work. The corpus, associated metadata, and code used to generate and analyze it are freely available.

\section{Methods}\label{sec:methods}

This research was deemed exempt from review by the Northwestern IRB. This corpus is derived exclusively from publicly available data, and no terms of service or electronic prevention methods were breached during data collection~\cite{boegershausen2021,krotov2018,krotov2020,gold2017}. Best-practice community standards were followed at all stages of work~\cite{rivers2014,zook2017,franzke2020}, such as making no attempt to gather additional information not already public, making no attempt to de-anonymize users, and being transparent with the data collection method. The released corpus is a heavily-processed derivative of the public HTML. Additional safeguards were applied to the released corpus to protect against future de-anonymization efforts, including the removal of non-essential data (e.g., user biographies), smoothing of demographic data (e.g., age binning), redaction of usernames, and redaction of names, emails, and URLs identified in posts.

\subsection{Data source}\label{sec:data-source}

\begin{figure}[t]
    \centering
    \includegraphics[width=\columnwidth]{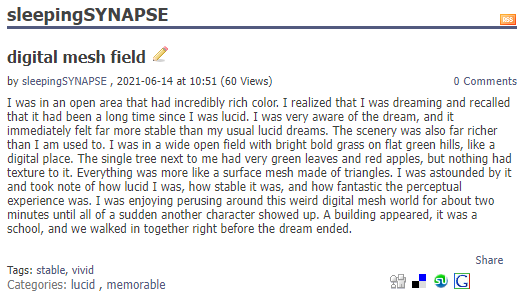}
    \caption{Example DreamViews dream journal entry. The categories labeling system, seen at the bottom, was used to extract labels for the corpus. This entry is from the author's journal.}\label{fig:journal-entry}
\end{figure}

DreamViews is a public and anonymous online forum dedicated to lucid dreaming. In addition to the discussion threads and user profiles typical of most forums, users can also share a public dream journal. An example dream journal entry is offered in Figure~\ref{fig:journal-entry}. Each journal entry includes a dream report (hereafter, ``post''), a title, and the option to attach ``category'' and ``tag'' labels that help identify features of the dream.

\par
The DreamViews labeling system plays a major role in the curation of this corpus because it allows for the isolation of certain types of dreams for study and comparison (similar to its online function of viewing dreams by label). When a user adds one or more category labels to their dream journal entry, their options are restricted to a small subset of terms provided by DreamViews (e.g., \emph{lucid}, \emph{non-lucid}, \emph{memorable}). In contrast, tag labels are unconstrained and thus consist of far more variety across participants (e.g., a barista who dreams about work a lot might be the only user with a \emph{coffee} tag), as well as varied spellings (e.g., \emph{non-lucid}, \emph{nonlucid}, and \emph{nld}). Only category labels were used to derive lucidity and nightmare labels, whereas future work might take advantage of the varied tag labels available in this corpus.

\subsection{Data collection}\label{sec:data-collection}

All public DreamViews dream journal entries between years 2010 and 2020 (inclusive) were scraped~\cite{speckmann2021,li2019,landers2016} using custom code on December 12\textsuperscript{th}, 2021, starting from the dream journal homepage\footnote{https://www.dreamviews.com/blogs} and crawling over all ``view all'' pages. Raw HTML was parsed to extract each journal entry's post (i.e., dream report text), title, tags, categories, timestamp, and user. The public profiles of dream journal users were scraped for self-reported demographic information (usernames were replaced with a random identifier). All text was converted to ASCII\footnote{https://github.com/avian2/unidecode}.

\subsection{Text processing}\label{sec:text-preprocessing}

Text was minimally cleaned with the aims of standardization, removing non-dream content, and removing potentially private or identifying information. Consecutive whitespaces were replaced with a single space, ampersands were replaced with text (\texttt{and}), common contractions were expanded, and sequences of four or more repeated characters~\cite{gray2020} were reduced to a single character. A regex pattern was used to replace emails and URLs with \texttt{<URL>}, and named entity recognition was used to identify and replace names with \texttt{<PERSON>}. Named entity recognition was implemented using \emph{spaCy}\footnote{https://spacy.io}, which typically outperforms related software~\cite{shelar2020,jiang2016}. Visual inspection revealed non-dream character sequences that were frequent and reliable across multiple posts, and these were removed using custom regex patterns (see online code for specifics).

\subsection{Data exclusion}\label{sec:data-exclusion}

Posts were removed if estimated as non-English\footnote{https://github.com/Mimino666/langdetect}, or containing \num{< 50} or $>$1k words (counted from cleaned text). In an effort to reduce single-user biases within the corpus, only the first 1k posts from a given user (that passed prior criteria) were kept. Visual inspection revealed, on minor occasion, systematic features of posts that indicated there was no dream content present (see online code for specifics). Such posts were removed.

\subsection{User-generated labels}\label{sec:automated-labeling}

The category labels attached to journal entries were used to identify the lucidity level of a post (\emph{lucid} or \emph{non-lucid}) and if the post was a \emph{nightmare} (see Figure~\ref{fig:journal-entry} and Section~\ref{sec:data-source}). For each journal entry where the user attached category labels, they were searched for relevant terms. To identify lucidity, terms ``lucid'' and ``non-lucid'' were searched; a post was identified as \emph{unspecified} if neither label was provided, \emph{ambiguous} if both labels were provided, and either \emph{lucid} or \emph{non-lucid} if one was provided without the other. Similarly, the term ``nightmare'' was searched, and a post was identified as a \emph{nightmare} if the term was present and \emph{non-nightmare} otherwise.

\begin{table}[t]
    \caption{Description of data files.}\label{table:datafiles}
    \begin{subtable}[t]{.48\textwidth}
        \caption{Columns of \texttt{dreamviews-posts.tsv}.}\label{table:datafile-posts}
        \small
        \begin{tabular}{rrl}
            \toprule
            column & dtype & description \\
            \midrule
            post\_id & str & unique randomized post identifier \\
            user\_id & str & unique randomized user identifier \\
            timestamp  & str & date and time of post (ISO 8601) \\
            nth\_post & int & user's nth post in corpus \\
            title & str & post title \\
            tags & str & tag labels of post \\
            categories & str & category labels of post \\
            lucidity & str & derived lucid label \\
            nightmare & bool & derived nightmare label \\
            wordcount & int & number of words in post \\
            post\_text & str & cleaned post text \\
            \bottomrule
        \end{tabular}
    \end{subtable}

    \bigskip
    \begin{subtable}[t]{.48\textwidth}
        \caption{Columns of \texttt{dreamviews-users.tsv}.}\label{table:datafile-users}
        \small
        \begin{tabular}{rrl}
            \toprule
            column & dtype & description \\
            \midrule
            user\_id & str & unique randomized user identifier \\
            gender  & str & reported gender \\
            age     & str & reported age bin \\
            country & str & reported country (ISO 3166-1 alpha-3) \\ % chktex 8
            \bottomrule
        \end{tabular}
    \end{subtable}
\end{table}

\begin{figure*}[t]
    \centering
    \includegraphics{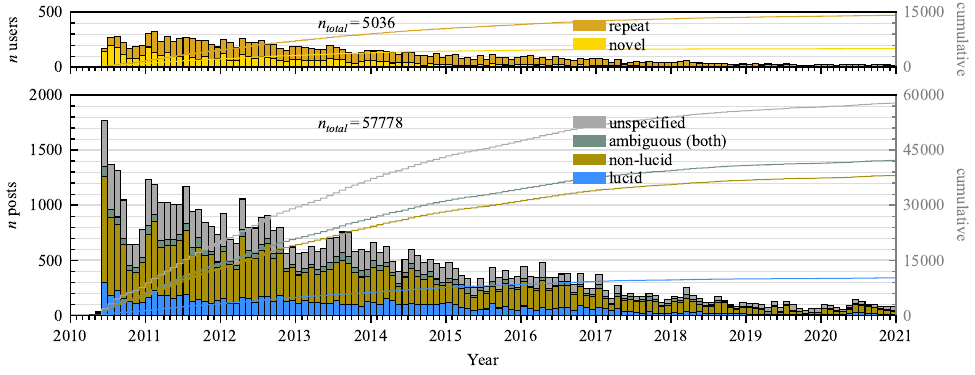}
    \caption{Total corpus size. Left axis values correspond to bars and right axis values correspond to cumulative distribution lines. Note that the final cumulative distributions endpoints provide a rough visualization of total corpus counts. The top panel shows the number of unique users that posted each month, where repeat/gold users are those that also posted in a prior month. The bottom panel shows the amount of unique posts each month, where colors indicate the lucidity label of posts. Note that the popular lucid dreaming film \emph{Inception} saw releases in July (USA theater) and December (USA home video) of 2010.}\label{fig:timecourse}
\end{figure*}

\begin{figure}[t]
    \centering
    \includegraphics{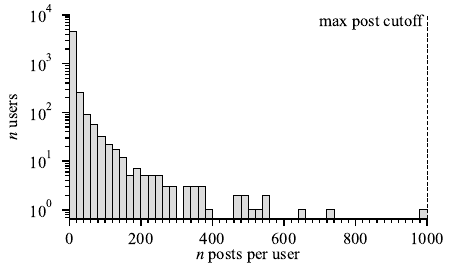}
    \caption{Number of posts per user.}\label{fig:usercount}
\end{figure}

\begin{figure*}
    \begin{subfigure}[t]{0.3\textwidth}
        \centering
        \includegraphics{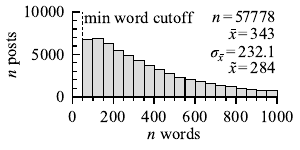}
        \caption{Word count per post. This histogram treats each dream report with equal weight, not accounting for repeated measurements within users.}\label{fig:wordcount1}
    \end{subfigure}
    \hfill
    \begin{subfigure}[t]{0.3\textwidth}
        \centering
        \includegraphics{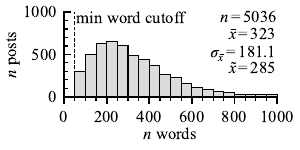}
        \caption{Word count per post, per user. Data in this histogram was first averaged to get a single word count per post for each individual user.}\label{fig:wordcount2}
    \end{subfigure}
    \hfill
    \begin{subfigure}[t]{0.3\textwidth}
        \centering
        \includegraphics{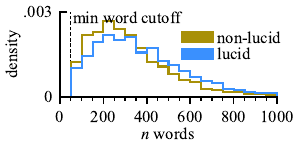}
        \caption{Word count per post, per user, by lucidity. Data in this histogram was first averaged to get a single word count per post for each individual user, separately for lucid and non-lucid posts. Lucid dream reports contained a higher word count than non-lucid dream reports.}\label{fig:wordcount3}
    \end{subfigure}
    \caption{Number of words per dream report. $n$: sample size, $\bar{x}$: mean, $\sigma_{\bar{x}}$: standard deviation, $\tilde{x}$: median}\label{fig:wordcount}
\end{figure*}

\newcolumntype{L}{>{\raggedright\arraybackslash}X}
\newcolumntype{R}{>{\raggedleft\arraybackslash}X}

\begin{table}[t]
    \caption{Top category and tag labels.}\label{table:toplabels}
    \begin{subtable}[t]{.25\textwidth}
        \caption{Top category labels.}\label{table:topcategories}
        \small
        \begin{tabular}{rp{1cm}}
            \toprule
            \bfseries category & \bfseries \makecell[rt]{$n$ posts\\$n$ users}
            \DTLforeach*{topcategories}{\cat=category, \nposts=n_posts, \nusers=n_users}{ % chktex 25
                \DTLiffirstrow{\\\midrule}
                \\\cat & \makecell[rt]{\num{\nposts}\\\num{\nusers}}
                \ifthenelse{\value{DTLrowi}=10}{\dtlbreak}{}
            }
            \\\bottomrule
        \end{tabular}
    \end{subtable}%
    \hfill
    \begin{subtable}[t]{.2\textwidth}
        \caption{Top tag labels.}\label{table:toptags}
        \small
        \begin{tabular}{rp{1cm}}
            \toprule
            \bfseries tag & \bfseries \makecell[rt]{$n$ posts\\$n$ users}
            \DTLforeach*{toptags}{\tag=tag, \nposts=n_posts, \nusers=n_users}{
                \DTLiffirstrow{\\\midrule}
                \\\tag & \makecell[rt]{\num{\nposts}\\\num{\nusers}}
                \ifthenelse{\value{DTLrowi}=10}{\dtlbreak}{}
            }
            \\\bottomrule
        \end{tabular}
    \end{subtable}
\end{table}

\begin{figure}[t]
    \centering
    \includegraphics{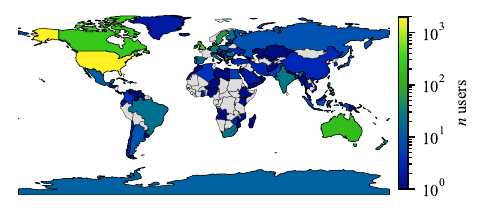}
    \caption{Reported user locations.}\label{fig:demographics-location}
\end{figure}

\begin{figure}[t]
    \centering
    \includegraphics{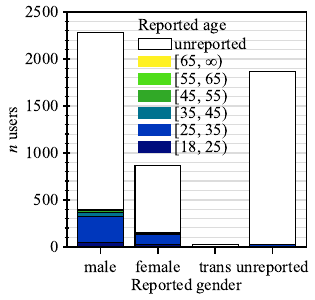}
    \caption{Reported user age and gender.}\label{fig:demographics-agegender}
\end{figure}

\subsection{Availability of materials}\label{sec:availability}

All code used to generate and analyze the current corpus is publicly available on GitHub\footnote{https://github.com/remrama/dreamviews}. The final corpus is available on Zenodo\footnote{https://doi.org/10.5281/zenodo.19161757} and abides by the FAIR principles of data management and stewardship~\cite{wilkinson2016}. See Table~\ref{table:datafiles} for a description of the data files.

\section{Data description}\label{sec:description}

\subsection{General characterization}\label{sec:general-characterization}

The final corpus contains \num{\totalposts} dream reports from \num{\totalusers} unique users (Figure~\ref{fig:timecourse}). Post frequency steadily declined over time, reducing to very little activity by 2021. Many users contributed multiple posts to the corpus, as indicated in the top panel of Figure~\ref{fig:timecourse} and in Figure~\ref{fig:usercount}. The most common category and tag labels are presented in Table~\ref{table:toplabels}.

\subsection{Demographics}\label{sec:demographics}

A variable amount of demographic information was self-reported in the public user profiles, including location (Figure~\ref{fig:demographics-location}) and age and gender (Figure~\ref{fig:demographics-agegender}). Of users in the final corpus, \num{\ngender} (\qty{\pctgender}{\percent}) provided a gender, \num{\nage} (\qty{\pctage}{\percent}) reported an age, \num{\ncountry} (\qty{\pctcountry}{\percent}) reported a location, and \num{\nall} (\qty{\pctall}{\percent}) users reported all three. Profiles suggest that users lean towards males in early adulthood from the United States. Non-English posts were removed during cleaning, which might have influenced location results.

\subsection{Word count}\label{sec:word-count}

Word count is a critical consideration when quantifying text and especially important when evaluating dream reports~\cite{urbina1981,schredl1999,schredl2011,rover2017,rosenlicht1994,antrobus1983}. Counts are presented in Figure~\ref{fig:wordcount} as the number of words per cleaned post (see Section~\ref{sec:text-preprocessing}). Word counts are presented in a few ways, emphasizing the importance of accounting for user dependencies and showing the difference between lucid and non-lucid posts.

\begin{figure}[t]
    \centering
    \includegraphics{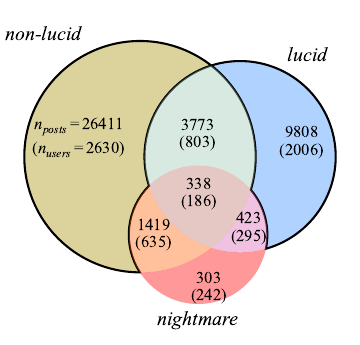}
    \caption{User-generated label frequencies, per post and per user. Within each section, the top number represents the number of posts and the bottom number represents the number of unique users that contribute to the post count. Non-lucid labels occur the most frequently, lucid labels second, and nightmare labels least. Venn sets are sized proportionally to user-normalized post frequencies.}\label{fig:categorycounts}
\end{figure}

\begin{figure}[t]
    \centering
    \includegraphics{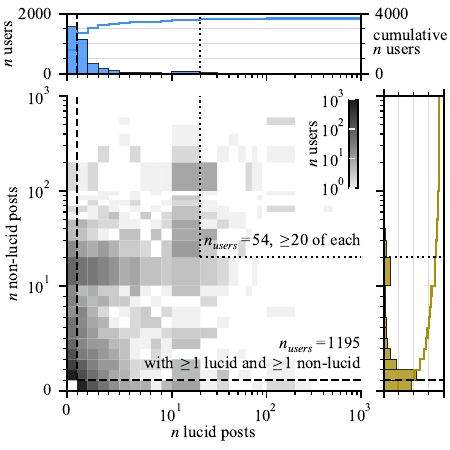}
    \caption{The frequency of users contributing lucid and non-lucid posts. The center panel highlights the number of users at each lucid and non-lucid post frequency. The dashed lines close a box around all post frequencies that include \num{>= 1} lucid and \num{>= 1} non-lucid, and the inner text provides the total number of users this is true for. The dotted lines close a box around all post frequencies that includes \num{>= 20} lucid and \num{>= 20} non-lucid, motivated by the finding that \num{20} dream reports provides a stable estimate of dream content within individuals~\cite{schredl1998}.
    Distributions in the marginal panels indicate the raw and cumulative number of users who contribute to either lucid or non-lucid post frequencies.}\label{fig:categorypairs}
\end{figure}

\subsection{Lucidity label frequency}\label{sec:lucidity-label-frequency}

Because the user-generated labeling of lucidity was not available for every post, it is important to understand what subset of the data \emph{does} include labels. The bottom panel of Figure~\ref{fig:timecourse} (cumulative distributions) provides a rough visualization of absolute and relative label frequencies. What that figure does not reveal is how many users contribute to each label (e.g., does one user contribute all the lucid posts?). Figure~\ref{fig:categorycounts} conveys this information, and also shows the overlap amongst labels more explicitly. This diagram shows there are roughly 14k posts labeled as lucid, with $\sim$4k of them also labeled as non-lucid and $\sim$500 of them also labeled as a nightmare. The overlap of lucid and non-lucid posts with nightmare labels shows the amount of data available for future analyses comparing lucid nightmares agains non-lucid nightmares. Another important consideration when choosing an analysis is the number of users contributing \emph{both} a lucid and a non-lucid post, as well as the total number of lucid and non-lucid posts each individual user contributes. These factors, presented in Figure~\ref{fig:categorypairs}, are important when considering statistical power and the reliability of dream reports~\cite{schredl1998}. The center shading represents the number of users who contribute each combination of lucid and non-lucid posts. The larger dashed line inset highlights the users who contributed one or more of each type of dream report (i.e., lucid and non-lucid). The smaller dashed subset indicates the same but for users with 20 or more of each post type. This visualization is useful in determining the amount of data for within-subjects lucid and non-lucid dream report comparisons. The marginal distributions isolate each dream type, showing the amount of users who contribute at each frequency level for lucid and non-lucid posts.

\section{Validation}\label{sec:validation}

The purpose of this section is to validate the user-generated labeling process described throughout prior sections. Before the labels can be used to make novel observations, they first need to be shown that they accurately represent the known features of the dream type they are supposed to represent (i.e., construct validity). Posts identified as lucid should ``look'' lucid~\cite{stumbrys2014,motarolim2013,bulkeley2018,voss2013,mallett2021}, and posts identified as a nightmare should look like nightmares~\cite{standards2010,bulkeley2018}.

\par
The validation steps proceed with increasing specificity in the following order. First, Section~\ref{sec:classifier} uses a classification approach to evaluate whether the language of lucid and non-lucid posts differ in a general sense. Next, Section~\ref{sec:wordshift} uses frequency-weighted measures to visualize word-level differences between lucid and non-lucid posts and assess their alignment with known lucid features (also for nightmares). Last, Section~\ref{sec:liwc} tests specific hypotheses about which psychological constructs appear more in lucid posts.

\subsection{General distinction}\label{sec:classifier}

This goal of this analysis was to determine whether lucid and non-lucid posts have different language use. A classifier was trained on word frequencies to identify a post as lucid or non-lucid.

\smallskip
\noindent
\textbf{Methods.} Post texts were tokenized, lemmatized, and lowercased using \emph{spaCy}. A token was removed if it contained a non-alphabetic character, was shorter than three characters, was a stop word or proper noun, or had a low probability of being in \emph{spaCy}'s vocabulary (to estimate mispellings or otherwise uncommon character sequences).

\par
Classification was implemented using \emph{scikit-learn}\footnote{https://scikit-learn.org}~\cite{pedregosa2011}. Each lemmatized post was transformed into a word-frequency vector. Very frequent (appearing in \qty{> 75}{\percent} of posts) or infrequent (appearing in \num{< 100} posts) words were removed, after which the top 5k words were kept as features. A linear support vector machine classifier was used with a regularization parameter of \num{1}. To avoid user-specific biases, a single post per user was randomly selected (irrespective of lucidity label) and then the more frequent class was downsampled to match the size of the smaller class. A stratified-shuffle-split cross-validation scheme was implemented, with the classifier trained on \qty{70}{\percent} of posts and tested on the other \qty{30}{\percent} at each of \num{5} folds. Performance was not optimized in any way (e.g., hyperparameter tuning).

\smallskip
\noindent
\textbf{Results.}
Lucid and non-lucid posts were easily distinguishable under this ``out-of-the-box'' classification model. Across cross-validation folds, classifier accuracy was \meansd{\clfaccm}{\clfaccsd} (M$\pm$SD, $\text{chance}=\num{0.50}$). Similar results were obtained with F1-score (\meansd{\clffonem}{\clffonesd}). This suggests that there are differences in language patterns between the lucid and non-lucid posts in this corpus and provide discriminant validity for the user-generated lucidity labels. Future work might optimize the present classification model and test its generalizability to novel and lab-based lucid dream reports.

\begin{figure}[t]
    \centering
    \includegraphics{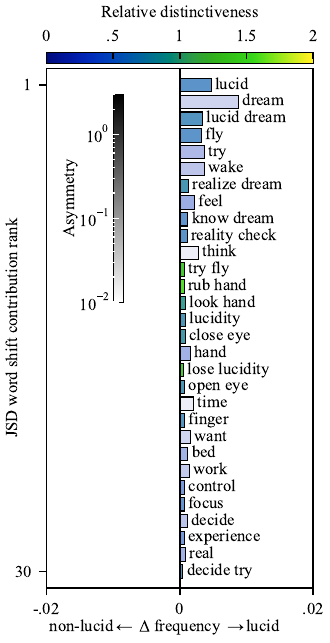}
    \caption{Lucidity JSD word shift. Words and phrases that most distinguish user-labeled lucid and non-lucid dream reports are consistent with known lucidity characteristics. These results offer content validity for user-generated lucidity labels.}\label{fig:wordshift-jsd}
\end{figure}

\begin{figure}[t]
    \centering
    \includegraphics{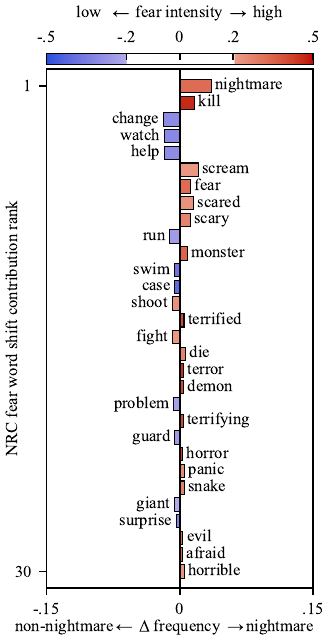}
    \caption{Nightmare fear word shift. Words and phrases that most distinguish user-labeled nightmares from all other dream reports are consistent with known nightmare characteristics. These results offer content validity for user-generated nightmare labels.}\label{fig:wordshift-fear}
\end{figure}

\begin{figure*}[t]
    \begin{subfigure}[t]{0.3\textwidth}
        \centering
        \includegraphics{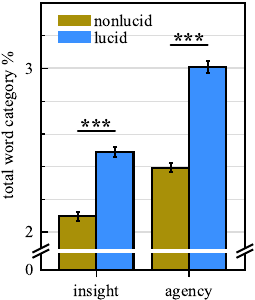}
        \caption{LIWC insight and agency language differences between user-labeled lucid and non-lucid dream reports. Lucid dream reports contained higher rates of both insight and agency than non-lucid dream reports. Error bars represent SEM, ***$p<.001$}\label{fig:liwc-categories}
    \end{subfigure}
    \hfill
    \begin{subfigure}[t]{0.3\textwidth}
        \centering
        \includegraphics{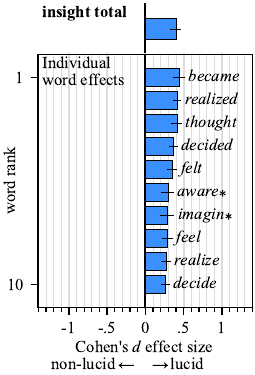}
        \caption{Word-level differences of insight language between user-labeled lucid and non-lucid dream reports. Insight was quantified as the percentage of words from LIWC's insight dictionary in each dream report. Error bars represent bootstrapped \qty{95}{\percent} confidence intervals. Asterisks indicate wildcard patterns.}\label{fig:liwc-words-insight}
    \end{subfigure}
    \hfill
    \begin{subfigure}[t]{0.3\textwidth}
        \centering
        \includegraphics{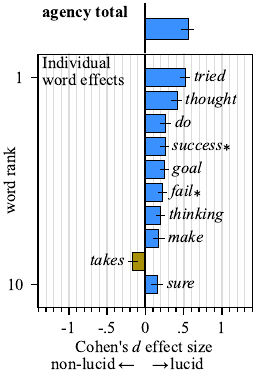}
        \caption{Word-level differences of agency language between user-labeled lucid and non-lucid dream reports. Agency was quantified as the percentage of words from LIWC's agency dictionary in each dream report. Error bars represent bootstrapped \qty{95}{\percent} confidence intervals. Asterisks indicate wildcard patterns.}\label{fig:liwc-words-agency}
    \end{subfigure}
    \caption{LIWC validation results. Dream reports labeled as lucid by the user contain a higher percentage of insight and agency dictionary words than those labeled as non-lucid dream reports. These results offer convergent validity for the user-labeled dream reports.}\label{fig:liwc}
\end{figure*}

\subsection{Word-level differentiation}\label{sec:wordshift}

The goal of this step was to validate labels by examining fine-grained differences between the language of posts and assess their consistency with predictable dream characteristics. Do lucid posts contain phrases common to lucid dream reports and culture, such as ``knew it was a dream'' and ``reality check''? Do nightmare posts contain phrases that represent fear, such as ``scared'' and ``monster''? These questions were answered using word shifts, an approach to comparing texts that takes traditional word-frequency differences and adds context-specific weights~\cite{gallagher2021}. Word shifts were used to identify the words (unigrams and bigrams) that contribute most to the separation between two texts (lucid vs.\ non-lucid, nightmare vs.\ non-nightmare), and the consistency of those words with prior literature was qualitatively assessed.

\smallskip
\noindent
\textbf{Methods.}
Word shift analyses were run on lemmatized posts (see Section~\ref{sec:classifier}) using \emph{Shifterator}\footnote{https://github.com/ryanjgallagher/shifterator}~\cite{gallagher2021} with common co-occuring unigrams first converted to bigrams using \emph{Gensim}\footnote{https://radimrehurek.com/gensim}~\cite{rehurek2010}. A detailed presentation of word shift methods are available elsewhere~\cite{gallagher2021}. The only additional step performed here was a normalization during initial calculation of word frequencies. Rather than summing word frequencies across all posts, user dependencies were accounted for by first dividing the word frequencies of each user by that user's number of posts (all within each corpus).

\par
To identify words that most distinguish lucid from non-lucid posts, the Jensen-Shannon divergence (JSD) was used to decompose the total distributional difference between the two corpora (lucid and non-lucid) into per-word contributions~\cite{gallagher2021,lin1991}. JSD captures how distinctively a word belongs to one corpus versus the other, making it well-suited to test the expectation that lucid posts would contain words that are rare in non-lucid contexts. When identifying top words, the overall JSD word shift score is determined by three factors: the difference in relative word frequency across lucid and non-lucid posts, relative word distinctiveness in relation to the average non-lucid word, and how asymmetrically the word is distributed between lucid and non-lucid posts.

\par
To identify words that most distinguish nightmare posts from non-nightmare (i.e., all other) posts, frequency differences were weighted by a word's fear-intensity score from the NRC Emotion Intensity Lexicon~\cite{mohammed2018}, a crowdsourced and freely available\footnote{http://saifmohammad.com/WebPages/AffectIntensity.htm} set of ratings for a variety of emotions. The motivation to use fear-weighted frequency differences was based on the clear expectation that nightmare posts should contain a higher frequency of words associated with fear. Only words with a fear intensity rating outside of range \numrange{-.2}{.2} (full range \numrange{-.5}{.5}) were included in this analysis to reduce the influence of neutral words. When identifying top words, the overall fear word shift score is determined by two factors: frequency difference between the two texts and the NRC fear intensity score.

\smallskip
\noindent
\textbf{Results.}
The JSD entropy word shift comparing lucid and non-lucid posts is presented in Figure~\ref{fig:wordshift-jsd}. Nearly all words that contribute to separating these texts are aligned with expected properties of lucid dream reports (content validity). First, the word ``lucid'' appears in multiple phrases that distinguish lucid and non-lucid posts (all at a higher frequency in lucid posts). Second, words relating to cognitive effort or deliberate action---such as ``think'', ``try'', ``control'', and ``want''---are distinguishing words that appear more frequently in lucid than non-lucid posts. Third, ``fly'' appears in multiple phrases that are more frequent in lucid than non-lucid posts, and flying is one of the most commonly reported lucid actions~\cite{stumbrys2014,motarolim2013,schadlich2012,barrett1991}. Fourth, many of the phrases that distinguish the corpora relate to lucid dream induction methods~\cite{stumbrys2012}, including ``reality check'', ``look hand'', and ``rub hand''. Fifth, distinguishing words such as ``feel'' and ``real'' are more frequent in lucid than non-lucid posts and might relate to an increased vividness in lucid dreams, as there are scattered reports that a subset of lucid dreams are particularly vivid~\cite{redditpaper}.

\par
The fear intensity word shift comparing nightmare and non-nightmare (i.e., unlabeled) posts is presented in Figure~\ref{fig:wordshift-fear}. Nightmare posts have a clear bias towards intense fear words. Words that appear more frequently in nightmare posts are also higher in fear intensity, including ``kill'', ``scream'', ``monster'', and ``panic''. In contrast, words low on fear intensity like ``swim'' and ``case'' occur less frequently in nightmare posts. The few fear words that \emph{do} appear more in non-nightmare posts---``fight'' and ``shoot''---are lower in fear intensity and consistent with the high rates of daily activities in dream content (e.g., sports or social quarrels)~\cite{schredl2010}.

\subsection{Cognitive language markers}\label{sec:liwc}

As a final validation step, lucid posts were tested for an increased presence of insight (i.e., awareness of the dream) and agency (i.e., control over dream content). Despite the difficulties in defining lucidity~\cite{mallett2021}, these two cognitive constructs are generally agreed to be common if not necessary components of a lucid dream~\cite{windt2007,voss2013,baird2019,stumbrys2014,mallett2021}. This perspective is well captured by the ``weak'' and ``strong'' criteria for lucidity~\cite{windt2007}; insight is the sole requirement for lucidity by the weak criterion, whereas it must be supplemented with agency to satisfy the strong criterion~\cite{windt2007}. While insight is fundamental to a lucid dream~\cite{voss2013,baird2019}, agency is another commonly associated feature~\cite{stumbrys2014,mallett2021,motarolim2013,voss2013,dresler2014}.

\smallskip
\noindent
\textbf{Methods.}
To capture the amount of insight and agency in language, validated word categories representative of each construct were used to generate word frequencies~\cite{tausczik2010}. This dictionary-based approach to text analysis has been applied extensively to dreams~\cite{bulkeley2018,hawkins2017,elce2021} and performs comparably with traditional dream content rating systems~\cite{zheng2021}. Dream insight was quantified in each post as the percentage of words that overlap with the \emph{insight} category from the LIWC2015 dictionary\footnote{https://liwc.app}~\cite{pennebaker2015} (e.g., ``think'', ``know''). Dream agency was quantified similarly using the \emph{agency} category from the Big Two dictionary\footnote{https://osf.io/p7fzb}~\cite{pietraszkiewicz2019} (e.g., ``doing'', ``making'', ``mastery'', ``overcome'', ``success'', ``fail''). Note that agency here is not intended to capture successful agency or dream control, but a sense of it. The language of failed attempts (e.g., ``fail'') would be expected to be higher when a sense of agency is present. Category-level and word-level percentages were calculated for each post using custom code and \emph{liwc-python}\footnote{https://github.com/chbrown/liwc-python}. Percentages were averaged within user, and only users with \num{>= 1} lucid posts and \num{>= 1} non-lucid posts were included (see dashed enclosing in Figure~\ref{fig:categorypairs}). LIWC insight and agency scores were compared between lucid and non-lucid dream reports. Wilcoxon signed-rank tests and common language effects sizes (CLES) were used to compare total insight and total agency between lucid and non-lucid posts. Cohen's $d$ effect sizes and their bootstrapped \qty{95}{\percent} confidence intervals (\num{2000} iterations) were calculated for overall insight and agency as well as each individual dictionary word. All statistics were calculated using \emph{Pingouin}\footnote{https://pingouin-stats.org}~\cite{vallat2018}.

\smallskip
\noindent
\textbf{Results.}
Category-level differences between lucid and non-lucid posts, for insight and agency, are presented in Figure~\ref{fig:liwc-categories}. Lucid posts contained far more insight (\wilcoxon{\insightW}{\insightP}{\insightCLES}) and agency (\wilcoxon{\agencyW}{\agencyP}{\agencyCLES}) words than non-lucid posts. These results suggest that lucid posts in this corpus have increased dream insight and agency, in line with prior reports of lucid dreams~\cite{windt2007,baird2019,voss2013,mallett2021}.

\par
Word-level effects comparing the individual words within insight and agency dictionaries are presented in Figures~\ref{fig:liwc-words-insight} and~\ref{fig:liwc-words-agency}. This analysis reveals the individual words that drive category-level differences between lucid and non-lucid posts for insight and agency. Ranked by absolute effect size, $d$, these results support the prior category-level findings in a few ways. First, they reveal that neither insight nor agency scores were driven by a small subset of words. With just one exception---``takes''---the top \num{10} words of each category are all in the expected direction, and with fairly even contributions. Second, they suggest that both categories are indeed picking up on the intended cognitive constructs of dream insight and agency. The largest insight effects seem very representative of lucid insight while dreaming (e.g., ``realize'', ``aware'') and easily fit into phrases describing the initial moment of lucidity while dreaming (e.g., I \emph{became} lucid; I \emph{realized} I was dreaming). Similarly, the strongest agency effects seem indicative of dream control (e.g., ``tried'', ``goal'', ``success''; I \emph{tried} to fly but \emph{failed}; I \emph{thought} about the \emph{goal} I set before bed). Together, these results serve as a strong validation of the lucidity labeling process, in that posts identified as lucid are consistent with known lucid dream phenomenology.

\section{Related datasets}\label{sec:related-datasets}

\smallskip
\noindent
\textbf{DreamBank.}
The DreamBank\footnote{https://dreambank.net} dataset is likely the first ``big'' dataset of dream reports to be released~\cite{domhoff2008} and has been analyzed extensively. With $>$22k dream reports (16k in English), DreamBank is comprised of many independent datasets that were collected under different contexts (e.g., surveys, individual dream journals). For example, within DreamBank is the classic Hall \& Van de Castle dataset, a collection of 1k dream reports that were used to develop an influential coding system and content norms~\cite{hall1966}. DreamBank's variety of studies has proven fruitful for novel discoveries. However, it also raises complicated dependencies across dream reports that need to be accounted for when viewing the corpus as a whole. Thus, it is common for researchers to extract subsets of DreamBank for analysis, such as a random collection~\cite{hendrickx2017} or a long dream series from an individual~\cite{fogli2020}. DreamBank is publicly viewable through an online interface and was recently released in a format more amenable to research\footnote{https://doi.org/10.5061/dryad.qbzkh18fr}~\cite{fogli2020}.

\smallskip
\noindent
\textbf{Sleep and Dream Database.}
The Sleep and Dream Database\footnote{https://sleepanddreamdatabase.org} (SDDb) is another big meta-dataset of dream reports~\cite{bulkeley2014} that has been analyzed extensively. At $\sim$30k dream reports, SDDb also contains rich cultural variety in the datasets it encompasses. The dream reports from SDDb are available through its online interface, where simple analyses can be run from a set of pre-determined options available in the browser. DreamBank and SDDb share many similarities in style and accessibility.

\smallskip
\noindent
\textbf{Digital trace dream studies.}
Individual studies have begun to reveal the usefulness of investigating dreams posted to online forums and social media. For example, dream-specific language markers were identified using 10k dream reports (subsetted from $>$119k posts and $>$73k users) from a short-lived social dream sharing site, DreamsCloud\footnote{https://tinyurl.com/2p8pm88n}~\cite{niederhoffer2017}. In another study, DreamBoard\footnote{https://dreamboard.com} posts were used to identify latent dream themes~\cite{mcnamara2019}, and a similar approach was taken to identify cross-cultural similarities in dream interpretation or ``meaning'' by scraping online dream dictionaries~\cite{varol2014}. Posts extracted from a variety of online dream journals (DreamJournal\footnote{http://dreamjournal.net}, InMyDream\footnote{https://tinyurl.com/5bjkv5a2}, YourDreamJournal\footnote{https://yourdreamjournal.wordpress.com}) were used to show that people take extra effort to normalize bizarre events when retelling their dreams~\cite{bardina2021}.

\par
Other lucid dreaming forums have also been utilized. An Italian lucid dreaming forum, SogniLucidi\footnote{https://www.sognilucidi.it/forum}, was used to identify gender~\cite{manna2019} and individual~\cite{manna2020} differences in dream reporting. A dataset of $>$200k posts from $>$15k users curated from DreamJournal was used to compare dreams and psychedelic experiences~\cite{sanz2018}. This DreamJournal dataset most closely aligns with the DreamViews corpus offered here. Whereas some DreamViews posts include a categorical lucidity identifier, many DreamJournal posts include a more continuous (Likert) lucidity rating.

\section{Limitations}\label{sec:limitations}

\textbf{Dream reports, not dreams.}
To study dreams, researchers are in the unfortunate position of having to rely on self-reports of dream experience (although see~\cite{konkoly2021,horikawa2013}), and this corpus is no exception. While the empirical study of subjective dream reports is a defendable position~\cite{windt2013,wamsley2013,solomonova2016,solomonova2014,gonzalez2021,ramm2018} and might be trainable~\cite{trnka2020,miyahara2020}, it is unrealistic to expect them to be completely veridical representations of dream experience~\cite{rosen2013,peels2016,boothe2015,bardina2021} or even contain strictly dream content~\cite{bardina2021}. Confabulations are likely one of many noise sources~\cite{putois2020} that contribute to dream reports, though the specifics of how dream reporting is biased is not clear. If confabulations of dream content during reporting are unique to individuals, this corpus or other big dream datasets might help to overcome this limitation (via brute force).

\smallskip
\noindent
\textbf{Non-binary lucidity.}
The labeling system here offers only a binary categorization of lucidity (i.e., lucid or non-lucid). Though a practical simplification, this can be problematic given that lucidity occurs in varying degrees~\cite{mallett2021}. A continuum view of lucidity might encompass dreams often referred to as pre-lucid or semi-lucid on the low end, and lucid dreams with full control on the high end~\cite{mallett2021,windt2007}. Future studies might explore this finer scale using the tag labels of the current corpus, which include \emph{pre-lucid} and \emph{semi-lucid} labels. Another interesting option would be to take an unsupervised clustering approach to reveal different levels of lucid phenomenology in this corpus.

\smallskip
\noindent
\textbf{Data bias and quality.}
Though digital trace data offers major benefits, they are not without cost~\cite{ruths2014,heng2018,rafaeli2019,lazer2020,boegershausen2021}. Considering the following limitations, any conclusions drawn about lucid dreaming from this corpus would be much stronger after replication in a separate corpus, either from another online source or a laboratory experiment~\cite{ruths2014}.

\par
First, as with laboratory studies, internet reports are not necessarily representative of truly ``natural'' or unbiased human behavior~\cite{ruths2014}. Context is likely to influence which dreams one selects for sharing, the extent of detail they choose to share, and the degree of confabulation. There might be a specific set of cultural and acceptable norms on DreamViews that impacts reporting and sharing~\cite{khalid2020,silva2009,ruths2014}.

\par
Second, demographic biases are likely to occur in any dataset compiled from social media (or worse, data is sometimes from bots). The current sample might be skewed towards American males and could potentially include more frequent video gamers than average, which would influence the amount of dream control expected in lucid dream reports~\cite{gackenbach2006,gackenbach2009,sestir2019}. In addition to cultural site preferences, internet access and necessary computer skills limit who is interested and able to use a given site~\cite{ruths2014,hargittai2020}. One speculative consideration is the potential bias towards certain lucid dreaming characteristics in the current sample, such as more (or less) natural lucid dreamers. Intriguingly, the number of participants that have experienced lucid dreaming in the general population is higher than those who have \emph{heard} about lucid dreaming~\cite{neuhausler2018}, suggesting that many consider lucid dreams ``normal'' and might not seek out related forums.

\par
Third, the general quality of internet data should be questioned~\cite{ruths2014,rafaeli2019}. Social media data made available through platform-specific APIs are controlled by the platforms, and have been shown in some cases to contain significant unadvertised biases~\cite{pfeffer2018,morstatter2017}. The current corpus did not use an API, though the web scraping approach contains more custom intervention steps (e.g., HTML parsing)~\cite{landers2016} and thus more room for error during data collection~\cite{braun2018,xu2020,boegershausen2021}. Similarly, datasets of this size are often too large for a complete manual quality assessment. These issues warrant caution in treating all text of the current corpus as dream content. A post might be entirely off topic (e.g., someone accidentally posting a forum topic to their dream journal), or a post that includes dream content might also contain non-dream expressions (e.g., introductory text prefacing the dream or normalizing devices buried within~\cite{bardina2021}). Another consideration is the case of multiple dreams within single dream reports~\cite{schredl2004}, which is not controlled for in this corpus. Based on cursory inspection, many consistent expressions of non-dream content are idiosyncratic among users, again stressing the importance of accounting for user dependencies in this corpus.

\section{Conclusion}\label{sec:conclusion}

This paper presented a new ``digital trace'' corpus of 55k dream reports collected from an online lucid dream journal. A population of 5k users contributed varying amounts of dream reports. A majority subset of 35k dream reports include labels---provided by the dreamer---that indicate whether the dream report describes a lucid (10k) or non-lucid (25k) dream. Analyses revealed that dream reports labeled as lucid were separable from those labeled as non-lucid based on their language use. Lucid-labeled posts contained more words commonly associated with lucid dreaming, and showed higher indications of insight and agency. A smaller subset (2k) of nightmare-labeled posts were also validated as containing higher levels of fear-related language than other posts. In summary, this large corpus of validated lucid and non-lucid dream reports opens up an opportunity for novel and reliable discoveries in dream science.

\section*{Declarations}

\textbf{Acknowledgements.}
RM would like to thank Ashwini Ashokkumar for invaluable support at all stages of this work.

\smallskip
\noindent
\textbf{Conflict of interest.} None declared.

\raggedbottom
\printbibliography

\end{document}